\documentclass[sigconf]{acmart}

\setcopyright{acmcopyright=0}
\copyrightyear{}
\acmYear{}
\acmDOI{}

\acmPrice{}
\acmISBN{}

\usepackage{times}
\usepackage{amsmath,amsfonts,amsthm,mathtools,graphicx,bm} 

\usepackage[T1]{fontenc}

\usepackage[utf8]{inputenc}

\usepackage{microtype}

\usepackage[ruled,vlined]{algorithm2e}
\usepackage{adjustbox, graphicx}

\usepackage{subfigure}
\newcommand{\bx}{\mathbf{x}}

\newcommand{\bt}{\mathbf{t}}

\newcommand{\bh}{\mathbf{h}}
\begin{document}

\title{Bi-Directional Recurrent Neural Ordinary Differential Equations  for  Social Media Text Classification}

\author{
    \textbf{Maunika Tamire,
    Srinivas Anumasa,
    P.K. Srijith}\\
    \textsuperscript{}Computer Science and Engineering\\
    Indian Institute of Technology Hyderabad, India\\
    cs18mds11026@iith.ac.in, cs16resch11004@iith.ac.in, srijith@cse.iith.ac.in\\
}

   
%

\renewcommand{\shortauthors}{}

\begin{abstract}
 Classification of posts in social media such as Twitter is difficult due to the noisy and short nature of texts. Sequence classification models based on  recurrent neural networks (RNN) are popular for classifying  posts that are sequential in nature. RNNs assume the hidden representation dynamics to evolve in a discrete manner and do not consider the exact time of the  posting. In this work, we propose to use recurrent neural ordinary differential equations (RNODE) for social media post classification which consider the time of posting  and allow the computation of hidden representation to evolve  in a time-sensitive continuous manner. In addition, we propose a novel model, Bi-directional RNODE (Bi-RNODE), which can consider the information flow in both the forward and backward directions of  posting times  to predict the post label. Our experiments demonstrate that  RNODE and Bi-RNODE are effective  for the problem of stance classification of rumours in social media. 

\end{abstract}





\maketitle

\section{Introduction}
Information disseminated in social media such as Twitter can be useful for addressing several real-world  problems like rumour detection, disaster management, and opinion mining.  Most of these problems involve classifying social media posts into different categories based on their textual content. For example, classifying the veracity of tweets as False, True, or unverified  allows one to debunk the rumours evolving in social media~\cite{zubiaga2018detection}.  However, social media text is extremely noisy with informal grammar, typographical errors, and  irregular vocabulary. In addition, the character limit (240 characters) imposed by social media such as Twitter make it even harder to perform text classification.  

Social media text classification, such as rumour stance classification\footnote{Rumour stance classification helps to identify the veracity of a rumour post by classifying the reply tweets into different stance classes such as {Support, Deny, Question, Comment}}~\cite{qazvinian2011rumor,zubiaga-etal-2016-stance,lukasik2016using} can be addressed effectively  using sequence labelling models  such as long short term memory (LSTM) networks~\cite{zubiaga-etal-2016-stance,augenstein-etal-2016-stance,kochkina2017turing,ZUBIAGA2018273,zubiaga2018detection,dey18,Liu2019TowardsEI,tian2020early}. 
Though they consider the sequential nature of tweets, they ignore the temporal aspects associated with the tweets. The time gap between tweets varies a lot and  LSTMs ignore this irregularity in tweet occurrences. They are discrete state space models where hidden representation changes from one tweet to another without considering the time difference between the tweets. Considering the exact times at which tweets occur can play an important role in determining the label. If the time gap between tweets is large, then the corresponding labels may not influence each other but can have a very high influence if they are closer.

We propose to use recurrent neural ordinary differential equations (RNODE)~\cite{rubanova2019latent} and developed a novel approach  bi-directional RNODE (Bi-RNODE), which can naturally consider the temporal information to perform  time sensitive classification of social media posts.  Neural ordinary differential equation (NODE)~\cite{node} is a continuous depth deep learning model  that performs transformation of feature vectors in a continuous manner using ordinary differential equation solvers. NODEs bring parameter efficiency and address model selection in deep learning to a great extent. 
Recurrent NODE~\cite{rubanova2019latent} extends NODE to time-series  where  hidden states associated with the elements in the sequence are assumed to evolve continuously over time.  They generalize RNNs to consider the temporal information present in the sequence data and allow the hidden representation to change according to this temporal information.

We propose RNODE to perform sequence labeling of posts occurring continuously over time in social media. It can consider the varying inter-arrival times in the posts and update the hidden representation according to it for classifying the posts. In addition, we propose a novel model, bi-directional RNODE (Bi-RNODE), which considers not only information from the past but also from the future in predicting the label of the post. Here, continuously evolving hidden representations in the forward and backward directions in time are combined and used to predict the post label. We show the effectiveness of the proposed models on the rumour stance classification problem in Twitter using the RumourEval-2019~\cite{RumourEval_2019_dataset} dataset. We found RNODE and Bi-RNODE can improve the social media text classification by effectively making use of the temporal information and is better than LSTMs and gated recurrent units (GRU)  with temporal features.

\section{Background}

\subsection{Problem Definition}
We consider the problem of classifying social media posts into different classes. Let us consider our data set $\mathcal{D}$ to be a collection of $N$ posts, $\mathcal{D} = \{p_i\}_{i=1}^N$. Each post $p_i$ is assumed to be a tuple containing details about the post such the textual content $\mathbf{x}_i$ (one can consider other features as well such as number of re-posts and reactions), time of the post $t_i$ and the label associated  with  the post $y_i$, thus  $p_i = \{(\mathbf{x}_i, t_i, y_i)\}$. Our aim is to develop a sequence classification model which consider  the temporal information $t_i$ along with $\bx_i$ for classifying a social media post. In particular, we consider the rumour stance classification problem in Twitter where one classify tweets into different classes such as Support, Query, Deny, and  Comment, thus $y_i$ $\in$ $\mathcal{Y}=\{Support, Query, Deny, Comment\}$. 

\subsection{Neural Ordinary Differential Equations}
Neural ordinary differential equations (NODE)\cite{node}  were introduced as a continuous depth alternative to Residual Networks (ResNets)\cite{he2016resnet}. 
ResNets uses skip connections to avoid vanishing gradient problems when networks grow deeper. Residual block output is computed as $\bh_{t+1}=\bh_t+f(\bh_t, \pmb{\theta}_t)$, where $f$ is a neural network  parameterized by $\theta_t$ involving stacked layers with non-linear activation functions and $\bh_t$ representing the hidden representation at depth $t$. This update is similar to a step in the Euler numerical technique used for solving ordinary differential equations (ODE) of the following form.
\begin{equation}
   \frac{d\bh(t)}{dt} = f(\bh(t), t, \pmb{\theta})
   \label{eqn:ode}
    \end{equation}
Sequence of residual block operations in ResNets can be seen as a solution to the ODE with $\bh(t)$ representing the hidden representation at any time $t$ and the ODE trajectories defined through the neural network $f$.  Consequently, NODEs can be interpreted as a continuous equivalent of ResNets modelling the evolution if hidden representations over time. 

    For solving ODE, one can use fixed step-size numerical techniques such as Euler, Runge-Kutta or adaptive step-size methods like Dopri5\cite{dormand1980family}. Solving an ODE requires one to specify an initial value ($\bh(0)$) and can compute the value at $t$ using an ODE solver $ODESolverCompute(f_{\pmb{{\theta}}} , \bh(0), 0, t)$. We can consider initial value $\bh(0)$ as input $\bx$ or a transformation of $\bx$ using a downsampling block. The ODE \eqref{eqn:ode} is solved until some end-time $T$ to obtain the final hidden representation $\bh(T)$. A fully connected neural network (FCNN) transforms the final representation $\bh(T)$ to the output $\hat{y}$. For classification problems cross-entropy loss is used to update the weights of NODE using back-propagation.  For NODE models, efficient back-propagation and gradient computations were proposed using adjoint sensitivity method~\cite{ACA12,node}.
\begin{figure}[t]
  \centering
\subfigure[RNODE architecture ]{  \includegraphics[scale=0.35]{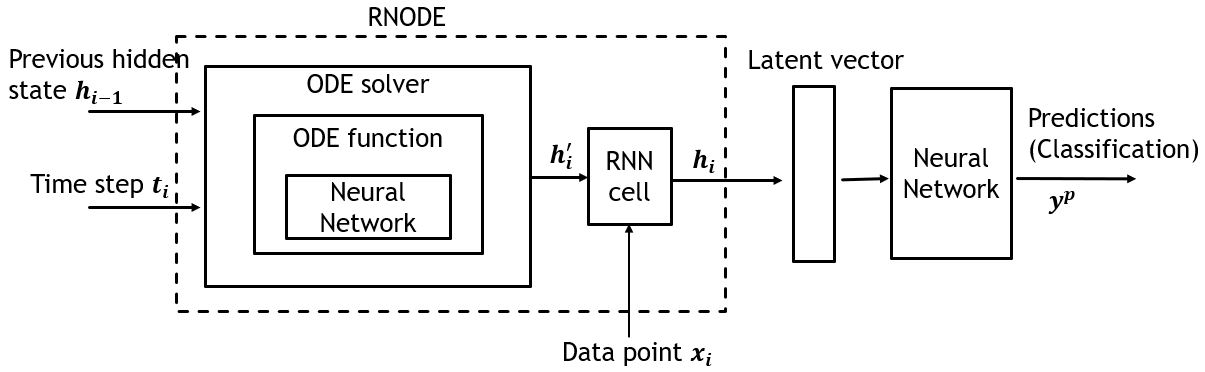}}
  
\subfigure[Bi-RNODE architecture]{   \includegraphics[scale=0.45]{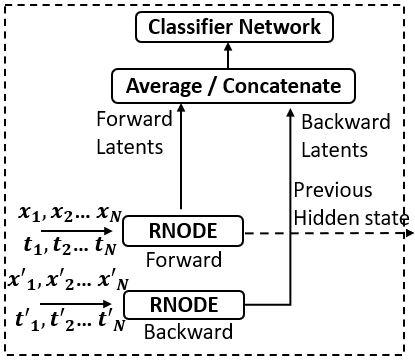}}
  \caption{Architecture details of RNODE and Bi-RNODE }
  \label{fig:RNODE-Bi-RNODE-detail}
\end{figure}

\section{Bi-Directional Recurrent NODE }

The popular techniques for sequence classification such as LSTMs consider the sequential nature of the data but ignores the temporal features associated with the data in its standard setting. The posts occur at irregular intervals of time, with more posts occurring at certain period. The influence of consecutive posts might depend on this time gap with the influence typically decreasing over time.  Instead of an LSTM model which perform single step transformation it will be beneficial to use a model where the number of transformations depend on the time gap. 

We propose to use recurrent neural ordinary differential equations (RNODE)~\cite{rubanova2019latent} to address the drawbacks of RNN based models in classifying irregularly occurring posts in social media. RNODE is developed for time-series data and can naturally consider the time associated with the posts make perform the transformations of the hidden representation to reflect the same.
In RNODE, the transformation of a hidden representation $\bh(t_{i-1})$ at time $t_{i-1}$ to $\bh(t_i)$ at time $t_i$ is governed by an ODE similar to \eqref{eqn:ode}, with $f$ being a neural network (NN) transformation. Unlike standard LSTMs where $\bh(t_i)$ is obtained from $\bh(t_{i-1})$ as a single NN transformation, RNODE first obtains a hidden representation $\bh'(t_i)$ as a solution to \eqref{eqn:ode} at time $t_i$ with initial value $\bh(t_{i-1})$. 
$$
\bh'(t_i) = \bh(t_{i-1}) + \int_{t_{i-1}}^{t_i} f(\bh(t_{i-1}), t_i - t_{i-1} , \pmb{\theta}).
$$
As this integral is intractable, RNODE uses a numerical technique (e.g., Euler method) to obtain the transformation. The number of update steps in the numerical technique is determined by the time gap $t_i - t_{i-1}$ between the consecutive posts.
$$\bh'(t_i) = ODESolverCompute(f_{\pmb{{\theta}}} , \bh(t_{i-1}), t_{i-1}, t_i).$$ 
The hidden representation $\bh'(t_i)$  and  input post $\bx_i$ at time $t_i$ are passed through neural network transformation (RNNCEll()) to obtain  final hidden representation $\bh(t_i)$, i.e., $\bh(t_i)$ = RNNCell($\bh'(t_i), \bx_i$).  The process is repeated for every element $(\bx_i, t_i)$ in the sequence. The hidden representations associated with the elements in the sequence are then passed to a neural network (NN()) to obtain the sequence of outputs corresponding to the post labels.   Figure \ref{fig:RNODE-Bi-RNODE-detail}(a) provides  the detailed architecture of the RNODE model.




%
Bi-directional RNNs~\cite{schuster1997bidirectional} such as Bi-LSTMS~\cite{graves13} were proven to be successful in many sequence labeling tasks in natural language processing such as POS tagging~\cite{Huang2015BidirectionalLM}. They use the information from the past and future to predict the label while standard LSTMs consider only from the past. We propose Bi-directional RNODE (Bi-RNODE), which uses the sequence of input observations from past and from the future to predict the post label at any time $t$. It assumes the hidden representation dynamics are influenced not only by the past posts but also by the futures posts. Unlike Bi-LSTMs, Bi-RNODE consider the exact time of the posts and their inter-arrival times in determining the transformations in the hidden representations.  Bi-RNODE consists of two RNODE blocks, one performing transformations in the forward direction (in the order of posting times) and the other in the backward direction (in the reverse order of posting times). The hidden representations  $H$ and  $H_b$ computed by forward and backward RNODE  respectively are aggregated either by concatenation or  averaging to obtain a final hidden representation and is passed through a NN to obtain the post labels.  Bi-RNODE is useful when a sequence of posts needs to be classified together, and can be restrictive for an online classification of individual posts. Algorithm \ref{algo:NODERNN} and Figure \ref{fig:RNODE-Bi-RNODE-detail}(b) provides an overview of Bi-RNODE  for post classification.  For Bi-RNODE, an extra neural network $f_{\pmb{\theta'}}()$ is required to compute hidden representations $H_b(t_i')$ in the backward direction. 
Training in Bi-RNODE is done in a similar manner to RNODE, with cross-entropy loss and back-propagation to estimate parameters.
\SetKwInput{KwInput}{Input} 
\SetKwInput{KwOutput}{Output} 
\begin{algorithm}[t]
\SetAlgoLined
\caption{Pseudo code for RNODE and Bi-RNODE approach to predict class labels. The input data points${(X, \bt)}$ where $X = \{\bx_i\}_{i=1}^N, \bt = \{t_i\}_{i=1}^N$ are sorted in increasing order of their timestamps.}
\label{algo:NODERNN}
Initialize: $\bh(0) = 0$, $t_{0} =0$, $H  = \{\}$  \\ 
\textbf{if} bidirectional:\\
\hspace{0.3cm} Set $X'$ to contain $X$ in reverse order. \\
\hspace{0.3cm} Set $\bt'$ to contain $\bt$ in  reverse order, where $t'_i = t_N - t_{N-i}$ \\ 
\hspace{0.3cm} $t'_{0} =0$ , $\bh_{b}(0) = 0$, $H_b$ $ = \{\}$
{
\\
    \For{$i\leftarrow 1$ \KwTo $N$}{\label{forins2}
    $\bh'(t_i)$ = ODESolverCompute($f_{\pmb{\theta}}$, $\bh(t_{i-1})$, $t_{i-1}$, $t_i$) \\
    $\bh(t_i)$ = RNNCell($\bh'(t_i)$ , $\bx_i$)\\
    $H = H \cup  \bh(t_i)$ \\
    \textbf{if} bidirectional:\\
    \hspace{0.3cm}$\bh'_b(t'_i)$  = ODESolverCompute($f_{\pmb{\theta'}}$), $\bh_{b}(t'_{i-1})$,$t'_{i-1}$, $t'_i$)\\
    \hspace{0.3cm}$\bh_{b}(t'_i)$ = RNNCell($\bh'_b(t'_i)$ , $\bx'_i$)\\
    \hspace{0.3cm}$H_b = H_b \cup  \bh_b(t'_i)$ \\
}
  \textbf{if}  bidirectional:\\
    \hspace{0.3cm} $H$ = aggregate($H$,$H_b$) \tcp{{\small concatenate or average}} 
    \textbf{return} NN$(H)$ \tcp{{\small return predicted post labels}}
}
\end{algorithm}

\section{Experiments}

To demonstrate the effectiveness of the proposed approaches, we consider the stance classification problem in Twitter and RumourEval-2019~\cite{RumourEval_2019_dataset} data set. This Twitter data set consists of rumours associated with eight events. Each event has collection of tweets labelled with one of the four labels - Support, Query, Deny and Comment. We picked four events Charliehebdo, Ferguson, Ottawashooting and Sydneysiege to conduct experiments.

{\bf Features} : For dataset preparation, each data point $\bx_i$ associated with a Tweet includes text embedding, retweet count, favourites count, punctuation features, sentiment polarity, negative and positive word count, presence of hashtags, user mentions, URLs, and entities etc. from the tweet information. Using pre-trained word2vec vectors~\footnote{Pre-trained vectors on Google News dataset: https://code.google.com/p/word2vec}, each word is represented as an embedding of size 15. The text embedding  of the tweet is obtained by concatenating the word embeddings. Each event data is split into train, validation, and test datasets with the ratio 60:20:20 in the order of time at which tweet occurred. Each tweet timestamp is converted to epoch time and Min-Max normalization is applied over the time stamps associated with each event to keep the duration of the event in the interval $[0,1]$.
\begin{figure*}[t]
  \begin{center}
      \subfigure[RNODE]{\includegraphics[scale=0.38]{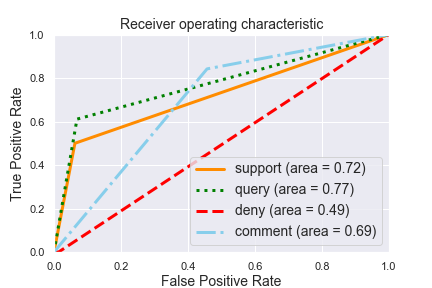}}
       \subfigure[LSTM]{\includegraphics[scale=0.38]{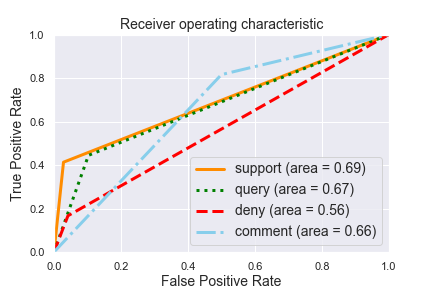}}
         \subfigure[GRU]{\includegraphics[scale=0.38]{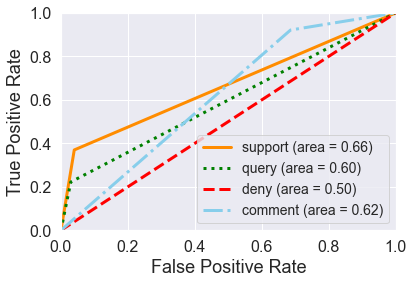}}
      \subfigure[Bi-RNODE]{\includegraphics[scale=0.38]{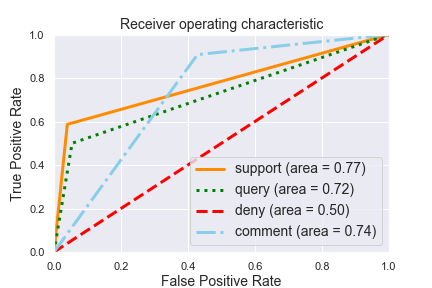}}
      \subfigure[Bi-LSTM]{\includegraphics[scale=0.38]{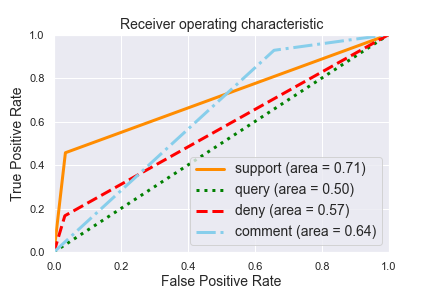}}
      \subfigure[Bi-GRU]{\includegraphics[scale=0.38]{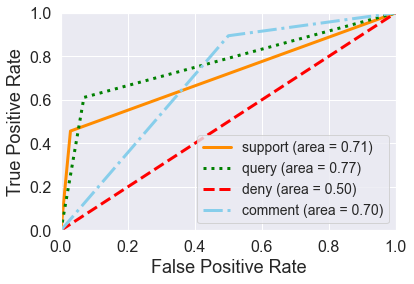}}
 \end{center}
 \vspace{-3mm}
 \caption{ROC curves of the models (a) RNODE (b) LSTM (c) GRU (d) BiRNODE (e) Bi-LSTM (f) Bi-GRU trained on sydneysiege event for seen event experiment }
 \label{fig:ROC-curves}
\end{figure*}

\subsection{Experimental setup}

In real time, new rumours arise and propagate at different time periods. Our experiments are conducted to predict stance of social media posts propagating in seen events as well as unseen events. Here are two experimental setups we conducted on the dataset.

\begin{itemize}
  \item \textbf{Seen Event}
Here we train, validate and test on tweets of same event. Each event data is split 60:20:20 ratio in sequence of time. This setup helps in predicting stance of unseen tweets of the same event.

\item \textbf{Unseen Event}: 
This setup helps  in evaluating performance on an unseen event and in training on a larger dataset. Here we consider training and validation on 3 events and testing on $4^{th}$ event. Last 20\% data of each of the training event is set aside for validation. During training, mini-batches are formed only from the posts in each event and are fed to the model in the order they appear in the event. 
\end{itemize}

\textbf{Baselines}:
We compared results of our proposed RNODE and Bi-RNODE models with RNN based baselines such LSTM~\cite{kochkina2017turing}, Bi-LSTM~\cite{augenstein-etal-2016-stance}, GRU~\cite{cho2014properties}, Bi-GRU, and Majority (labelling most frequent class) baseline models. We also use  a variant of  LSTM  baseline considering temporal information~\cite{ZUBIAGA2018273}, LSTM-timeGap where the  timegap of consecutive data points is included as part of the input data.

\textbf{Evaluation Metrics}:
We consider the standard evaluation metrics   such as precision, recall, F1 and in addition the  AUC score to account for the  data imbalance. We consider a weighted average of the evaluation metrics to compare the performance of models.


\begin{table*}[t]
\caption{Performance of all the models on RumourEval-2019~\cite{RumourEval_2019_dataset} dataset. First and second rows of each model represents \textit{seen event} and \textit{unseen event} experiment results respectively. 
}
\label{tab:resultsIndividual}
\begin{tabular}{|l|l|l|l|l|l|l|l|l|l|l|l|l|l|l|l|l|l|}
\hline
{\small \textbf{Model}} & \multicolumn{4}{c|}{Charliehebdo} & \multicolumn{4}{c|}{Ferguson} & \multicolumn{4}{c|}{Ottawashooting} & \multicolumn{4}{c|}{Sydneysiege}  \\
\hline
               & {\small \textbf{AUC}} & {\small \textbf{F1}} & {\small \textbf{Recall}} & {\small \textbf{Preci-}}  & {\small \textbf{AUC}} & {\small \textbf{F1}} & {\small \textbf{Recall}} & {\small \textbf{Preci-}} & {\small \textbf{AUC}} & {\small \textbf{F1}} & {\small \textbf{Recall}} & {\small \textbf{Preci-}} &{\small  \textbf{AUC}} & {\small \textbf{F1}} & {\small \textbf{Recall}} & {\small \textbf{Preci-}}\\
               & & & & {\small \textbf{sion}} & & & & {\small \textbf{sion}} & & & & {\small \textbf{sion}}& & & & {\small \textbf{sion}} \\
\hline
 {\small RNODE}    & {\small 0.665} & {\small0.653} & {\small0.674} & {\small\textbf{0.658}} & {\small\textbf{0.600}} & {\small0.591} & {\small0.659} & {\small0.598} & {\small0.638} & {\small0.654} & {\small\textbf{0.692}} & {\small\textbf{0.670}} & {\small0.699} & {\small0.722} & {\small0.730} & {\small0.724} \\
   &  {\small 0.638} & {\small0.672} & {\small0.700} & {\small\textbf{0.721}} & {\small\textbf{0.618}} & {\small0.632} & {\small0.677} & {\small\textbf{0.640}} & {\small\textbf{0.659}} & {\small\textbf{0.651}} & {\small\textbf{0.703}} & {\small0.642} & {\small\textbf{0.661}} & {\small 0.662} & {\small 0.704} & {\small0.638} \\
\hline
 {\small Bi-RNODE} & {\small\textbf{0.696}} & {\small\textbf{0.659}} & {\small\textbf{0.693}} & {\small0.629} & {\small0.595} & {\small 0.599 } & {\small\textbf{0.673}} & {\small\textbf{0.641}} & {\small\textbf{0.669}} & {\small\textbf{0.667}} & {\small\textbf{0.692}} & {\small0.658}  & {\small\textbf{0.739}} & {\small\textbf{0.769}} & {\small\textbf{0.784}} & {\small\textbf{0.763}} \\
& {\small 0.651} & {\small\textbf{0.697}} & {\small\textbf{0.737}} & {\small0.690} & {\small0.615} & {\small\textbf{0.643}} & {\small\textbf{0.695}} & {\small0.635} & {\small0.652} & {\small0.624} & {\small0.662} & {\small0.618} & {\small0.650} & {\small0.650} & {\small0.669} & {\small0.653} \\
 \hline
 {\small Bi-LSTM}  & {\small 0.628} & {\small 0.625} & {\small 0.679} & {\small 0.609} & {\small 0.563} & {\small 0.599} & {\small 0.650} & {\small 0.614} & {\small 0.622} & {\small 0.627} & {\small 0.654} & {\small 0.622} & {\small 0.648} & {\small 0.701} & {\small 0.716} & {\small 0.721} \\
 
            & {\small 0.662} & {\small0.690} & {\small0.717} & {\small0.671} & {\small0.603} & {\small0.623} & {\small0.667} & {\small0.600} & {\small0.650} & {\small0.637} & {\small0.686} & {\small0.622} & {\small0.652} & {\small0.655} & {\small0.680} & {\small0.652} \\
 \hline
 {\small Bi-GRU} & {\small 0.654} & {\small 0.643} & {\small 0.660} & {\small 0.641} & {\small 0.588} & {\small 0.571} & {\small 0.631} & {\small 0.625} & {\small 0.640} & {\small 0.651} & {\small 0.686} & {\small 0.644} & {\small 0.701} & {\small 0.739} & {\small 0.757} & {\small 0.748} \\ 
 & {\small 0.656} & {\small 0.690} & {\small 0.724} & {\small 0.682} & {\small 0.613} & {\small 0.634} & {\small 0.678} & {\small 0.611} & {\small 0.648} & {\small 0.636} & {\small 0.683} & {\small 0.610} & {\small 0.653} & {\small 0.655} & {\small 0.690} & {\small 0.680} \\
 \hline
 {\small LSTM}  & {\small0.625} & {\small0.600} & {\small0.637} & {\small0.637} & {\small0.567} & {\small\textbf{0.602}} & {\small0.650} & {\small0.611} & {\small0.605} & {\small0.609} & {\small0.635} & {\small0.603} & {\small0.634} & {\small0.689} & {\small0.703} & {\small0.695} \\
        & {\small 0.645} & {\small 0.690} & {\small0.728} & {\small0.686} & {\small0.602} & {\small0.611} & {\small0.631} & {\small0.603} & {\small0.630} & {\small0.626} & {\small0.680} & {\small0.627} & {\small0.643} & {\small0.644} & {\small0.669} & {\small0.641}\\
 \hline
 {\small GRU} & {\small 0.616} & {\small 0.610} & {\small 0.647} & {\small 0.623} & {\small 0.578} & {\small 0.588} & {\small 0.664} & {\small 0.631} & {\small 0.591} & {\small 0.539} & {\small 0.513} & {\small 0.574} & {\small 0.623} & {\small 0.688} & {\small 0.725} & {\small 0.689} \\ 
 & {\small \textbf{0.682}} & {\small 0.695} & {\small 0.713} & {\small 0.686} & {\small 0.614} & {\small 0.640} & {\small 0.687} & {\small 0.623} & {\small 0.638} & {\small 0.632} & {\small 0.683} & {\small 0.618} & {\small 0.654} & {\small \textbf{0.665}} & {\small \textbf{0.711}} & {\small \textbf{0.659}} \\
 \hline
 {\small LSTM-}  & {\small 0.638} & {\small0.631} & {\small0.679} & {\small0.605} & {\small0.565} & {\small0.581} & {\small0.627} & {\small0.590} & {\small0.625} & {\small0.640} & {\small0.679} & {\small0.650} & {\small0.656} & {\small0.667} & {\small0.667} & {\small0.671}  \\
 {\small timeGap} 
  & {\small0.652} & {\small0.695} & {\small0.732} & {\small0.696} & {\small0.604} & {\small0.625} & {\small0.673} & {\small0.633} & {\small0.638} & {\small0.638} & {\small0.683} & {\small\textbf{0.651}}  & {\small0.632} & {\small0.649} & {\small0.698} & {\small0.655}\\
 \hline
 {\small Majority} & {\small0.500} & {\small0.456} & {\small0.605} & {\small0.366}   & {\small0.500} & {\small0.518} & {\small0.654} & {\small0.428} & {\small0.500} & {\small0.485} & {\small0.628} & {\small0.395} & {\small0.500} & {\small0.485} & {\small0.628} & {\small0.395} \\
  & {\small 0.500} & {\small0.542} & {\small0.673} & {\small0.453} & {\small0.500} & {\small0.528} & {\small0.662} & {\small0.439} & {\small0.500} & {\small0.467} & {\small0.614} & {\small0.377} & {\small0.500} & {\small0.490} & {\small0.632} & {\small0.400} \\
 \hline
\end{tabular}
\end{table*}

\textbf{Hyperparameters}:
All the models are trained for 50 epochs with 0.01 learning rate, Adam optimizer, dropout(0.2) regularizer, batchsize of 50 and cross entropy loss function. 
Different hyperparameters like neural network layers (1, 2),  hidden representation  sizes (64,128), numerical methods (Euler, RK4, Dopri5 for RNODE and Bi-RNODE) and aggregation strategy (concatenation or averaging for Bi-LSTM and Bi-RNODE) are used for all the models and the best configuration is selected from the validation data for different experimental setups and train/test data splits.

\begin{figure}[t]
  \begin{center}
      \subfigure[RNODE]{\includegraphics[scale=0.1652]{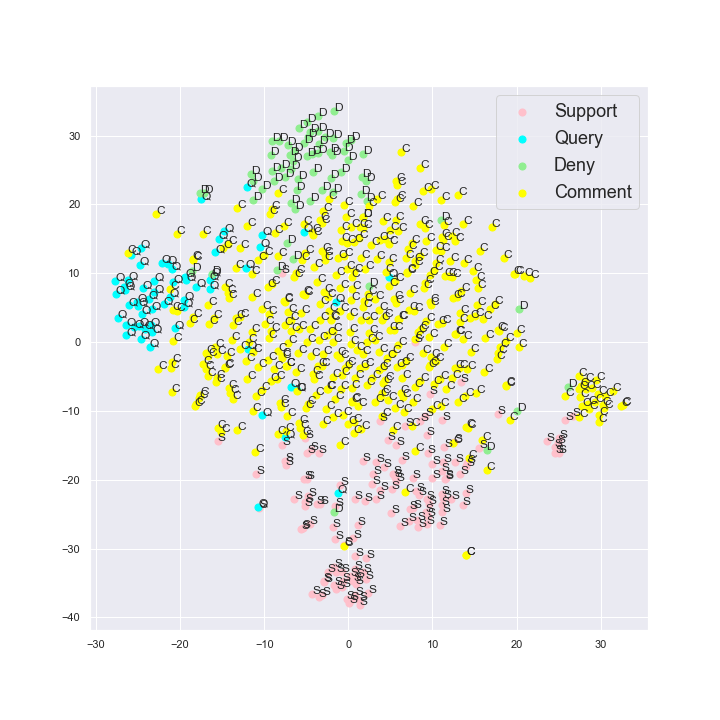}}
      \subfigure[Bi-RNODE]{\includegraphics[scale=0.1652]{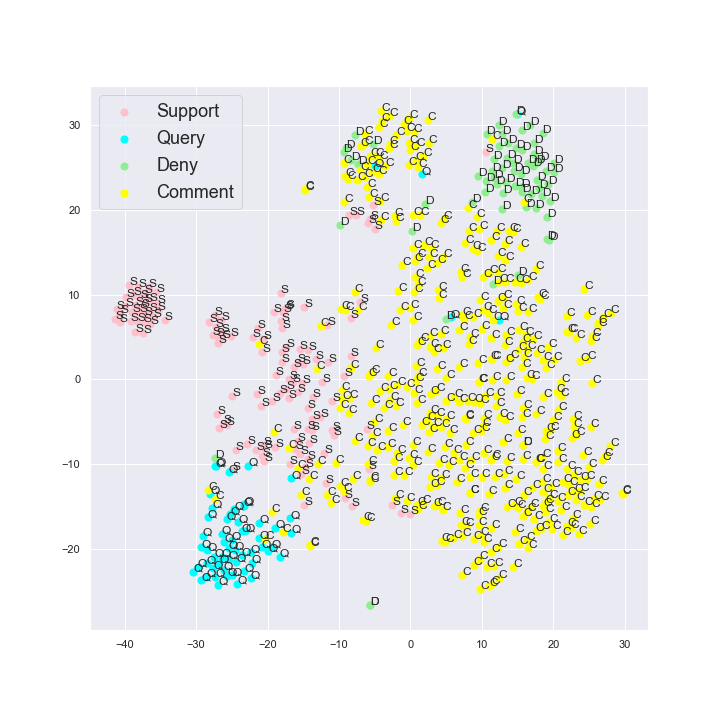}}
      \caption{t-SNE plot of (a) RNODE and (b) Bi-RNODE latent representations for the Sydneysiege event }
  \label{fig:tsne}
  \end{center}
  \end{figure}
\subsection{Results and Analysis}
The results of  \textit{seen event} and \textit{unseen event} experiment setup can be found in Table~\ref{tab:resultsIndividual}, where the  first and second rows for each model provides results on \textit{seen event} and \textit{unseen event} respectively. 
We can observe from Table~\ref{tab:resultsIndividual} that for both  \textit{seen event} and  \textit{unseen event} experiment setup, our proposed RNODE and Bi-RNODE models outperformed baseline models for all the four events. For the \textit{seen event} setup, Bi-RNODE gives the best result out-performing other models for most of the data sets and measures.  While for unseen event setup, RNODE and Bi-RNODE models gave better results when compared to baseline models except for Charliehebdo event. Bi-RNODE results are better than RNODE for Charliehebdo and Ferguson, while it is close to RNODE for Ottawashooting and Sydneysiege. Under seen event experiment on syndneysiege event, we plot the ROC curve for all the models in Figure~\ref{fig:ROC-curves}. We can observe that AUC for Figures \ref{fig:ROC-curves}(a) and \ref{fig:ROC-curves}(d) corresponding to RNODE and Bi-RNODE respectively are higher  than LSTM, GRU, Bi-LSTM , and Bi-GRU.

The proposed models are computationally and parametrically efficient where RNODE ($0.22$M,in Millions) and Bi-RNODE ($0.33$M) models required less parameters when compared to LSTM ($1.70$M) and Bi-LSTMS ($3.40$M) models. Visualization of latent hidden state representations of the proposed models using t-SNE plot (Figure \ref{fig:tsne}(a) and \ref{fig:tsne}(b)) shows that they are capable of separating data points from different classes into different groups. The proposed models are learning hidden representations well with Bi-RNODE learning a better representation than RNODE. 

\section{Conclusion and Future Work}
We proposed RNODE and Bi-RNODE models for sequence classification of social media posts which  naturally consider the temporal information and  use it to model the dynamics of hidden representations using an ODE. 
This makes them more effective than LSTMs for social media  where posts occur at irregular time intervals.  The experimental results on the rumour stance classification problem in Twitter supports the superior capability of the RNODE and Bi-RNODE in performing tweet classification. As a future work, we would like to further improve the sequence modelling capability of the proposed models by combining them with conditional random fields. 

\bibliographystyle{ACM-Reference-Format}
\bibliography{sample-base}


\begin{thebibliography}{20}


\ifx \showCODEN    \undefined \def \showCODEN     #1{\unskip}     \fi
\ifx \showDOI      \undefined \def \showDOI       #1{#1}\fi
\ifx \showISBNx    \undefined \def \showISBNx     #1{\unskip}     \fi
\ifx \showISBNxiii \undefined \def \showISBNxiii  #1{\unskip}     \fi
\ifx \showISSN     \undefined \def \showISSN      #1{\unskip}     \fi
\ifx \showLCCN     \undefined \def \showLCCN      #1{\unskip}     \fi
\ifx \shownote     \undefined \def \shownote      #1{#1}          \fi
\ifx \showarticletitle \undefined \def \showarticletitle #1{#1}   \fi
\ifx \showURL      \undefined \def \showURL       {\relax}        \fi
\providecommand\bibfield[2]{#2}
\providecommand\bibinfo[2]{#2}
\providecommand\natexlab[1]{#1}
\providecommand\showeprint[2][]{arXiv:#2}

\bibitem[\protect\citeauthoryear{Augenstein, Rockt{\"a}schel, Vlachos, and
  Bontcheva}{Augenstein et~al\mbox{.}}{2016}]%
        {augenstein-etal-2016-stance}
\bibfield{author}{\bibinfo{person}{Isabelle Augenstein}, \bibinfo{person}{Tim
  Rockt{\"a}schel}, \bibinfo{person}{Andreas Vlachos}, {and}
  \bibinfo{person}{Kalina Bontcheva}.} \bibinfo{year}{2016}\natexlab{}.
\newblock \showarticletitle{Stance Detection with Bidirectional Conditional
  Encoding}. In \bibinfo{booktitle}{\emph{Proceedings of the 2016 Conference on
  Empirical Methods in Natural Language Processing}}.
  \bibinfo{pages}{876--885}.
\newblock


\bibitem[\protect\citeauthoryear{Chen, Rubanova, Bettencourt, and
  Duvenaud}{Chen et~al\mbox{.}}{2018}]%
        {node}
\bibfield{author}{\bibinfo{person}{Ricky~TQ Chen}, \bibinfo{person}{Yulia
  Rubanova}, \bibinfo{person}{Jesse Bettencourt}, {and}
  \bibinfo{person}{David~K Duvenaud}.} \bibinfo{year}{2018}\natexlab{}.
\newblock \showarticletitle{Neural ordinary differential equations}. In
  \bibinfo{booktitle}{\emph{Advances in neural information processing
  systems}}. \bibinfo{pages}{6571--6583}.
\newblock


\bibitem[\protect\citeauthoryear{Cho, van Merri{\"e}nboer, Bahdanau, and
  Bengio}{Cho et~al\mbox{.}}{2014}]%
        {cho2014properties}
\bibfield{author}{\bibinfo{person}{Kyunghyun Cho}, \bibinfo{person}{Bart van
  Merri{\"e}nboer}, \bibinfo{person}{Dzmitry Bahdanau}, {and}
  \bibinfo{person}{Yoshua Bengio}.} \bibinfo{year}{2014}\natexlab{}.
\newblock \showarticletitle{On the Properties of Neural Machine Translation:
  Encoder--Decoder Approaches}. In \bibinfo{booktitle}{\emph{Proceedings of
  SSST-8, Eighth Workshop on Syntax, Semantics and Structure in Statistical
  Translation}}. \bibinfo{pages}{103--111}.
\newblock


\bibitem[\protect\citeauthoryear{Derczynski, Gorrell, Zubiaga, Aker, Bontcheva,
  Liakata, and Kochkina}{Derczynski et~al\mbox{.}}{2019}]%
        {RumourEval_2019_dataset}
\bibfield{author}{\bibinfo{person}{Leon Derczynski}, \bibinfo{person}{Genevieve
  Gorrell}, \bibinfo{person}{Arkaitz Zubiaga}, \bibinfo{person}{Ahmet Aker},
  \bibinfo{person}{Kalina Bontcheva}, \bibinfo{person}{Maria Liakata}, {and}
  \bibinfo{person}{Elena Kochkina}.} \bibinfo{year}{2019}\natexlab{}.
\newblock \bibinfo{title}{RumourEval 2019 data}.
\newblock
\newblock
\urldef\tempurl%
\url{https://doi.org/10.6084/M9.FIGSHARE.8845580.V1}
\showDOI{\tempurl}


\bibitem[\protect\citeauthoryear{Dey, Shrivastava, and Kaushik}{Dey
  et~al\mbox{.}}{2018}]%
        {dey18}
\bibfield{author}{\bibinfo{person}{Kuntal Dey}, \bibinfo{person}{Ritvik
  Shrivastava}, {and} \bibinfo{person}{Saroj Kaushik}.}
  \bibinfo{year}{2018}\natexlab{}.
\newblock \showarticletitle{Topical Stance Detection for Twitter: A Two-Phase
  LSTM Model Using Attention}. In \bibinfo{booktitle}{\emph{Advances in
  Information Retrieval}}. \bibinfo{pages}{529--536}.
\newblock


\bibitem[\protect\citeauthoryear{Dormand and Prince}{Dormand and
  Prince}{1980}]%
        {dormand1980family}
\bibfield{author}{\bibinfo{person}{John~R Dormand} {and}
  \bibinfo{person}{Peter~J Prince}.} \bibinfo{year}{1980}\natexlab{}.
\newblock \showarticletitle{A family of embedded Runge-Kutta formulae}.
\newblock \bibinfo{journal}{\emph{Journal of computational and applied
  mathematics}} \bibinfo{volume}{6}, \bibinfo{number}{1}
  (\bibinfo{year}{1980}), \bibinfo{pages}{19--26}.
\newblock


\bibitem[\protect\citeauthoryear{Graves, Mohamed, and Hinton}{Graves
  et~al\mbox{.}}{2013}]%
        {graves13}
\bibfield{author}{\bibinfo{person}{Alex Graves}, \bibinfo{person}{Abdel-rahman
  Mohamed}, {and} \bibinfo{person}{Geoffrey Hinton}.}
  \bibinfo{year}{2013}\natexlab{}.
\newblock \showarticletitle{Speech recognition with deep recurrent neural
  networks}. In \bibinfo{booktitle}{\emph{2013 IEEE International Conference on
  Acoustics, Speech and Signal Processing}}. \bibinfo{pages}{6645--6649}.
\newblock


\bibitem[\protect\citeauthoryear{He, Zhang, Ren, and Sun}{He
  et~al\mbox{.}}{2016}]%
        {he2016resnet}
\bibfield{author}{\bibinfo{person}{Kaiming He}, \bibinfo{person}{Xiangyu
  Zhang}, \bibinfo{person}{Shaoqing Ren}, {and} \bibinfo{person}{Jian Sun}.}
  \bibinfo{year}{2016}\natexlab{}.
\newblock \showarticletitle{Deep residual learning for image recognition}. In
  \bibinfo{booktitle}{\emph{Proceedings of the IEEE conference on computer
  vision and pattern recognition}}. \bibinfo{pages}{770--778}.
\newblock


\bibitem[\protect\citeauthoryear{Huang, Xu, and Yu}{Huang
  et~al\mbox{.}}{2015}]%
        {Huang2015BidirectionalLM}
\bibfield{author}{\bibinfo{person}{Zhiheng Huang}, \bibinfo{person}{W. Xu},
  {and} \bibinfo{person}{Kai Yu}.} \bibinfo{year}{2015}\natexlab{}.
\newblock \showarticletitle{Bidirectional {LSTM-CRF} Models for Sequence
  Tagging}.
\newblock \bibinfo{journal}{\emph{ArXiv abs/1508.01991}}
  (\bibinfo{year}{2015}).
\newblock


\bibitem[\protect\citeauthoryear{Kochkina, Liakata, and Augenstein}{Kochkina
  et~al\mbox{.}}{2017}]%
        {kochkina2017turing}
\bibfield{author}{\bibinfo{person}{Elena Kochkina}, \bibinfo{person}{Maria
  Liakata}, {and} \bibinfo{person}{Isabelle Augenstein}.}
  \bibinfo{year}{2017}\natexlab{}.
\newblock \showarticletitle{{T}uring at {S}em{E}val-2017 Task 8: Sequential
  Approach to Rumour Stance Classification with Branch-{LSTM}}. In
  \bibinfo{booktitle}{\emph{Proceedings of the 11th International Workshop on
  Semantic Evaluation ({S}em{E}val-2017)}}. \bibinfo{pages}{475--480}.
\newblock


\bibitem[\protect\citeauthoryear{Liu, Jin, and Shen}{Liu et~al\mbox{.}}{2019}]%
        {Liu2019TowardsEI}
\bibfield{author}{\bibinfo{person}{Y. Liu}, \bibinfo{person}{X. Jin}, {and}
  \bibinfo{person}{H. Shen}.} \bibinfo{year}{2019}\natexlab{}.
\newblock \showarticletitle{Towards early identification of online rumors based
  on long short-term memory networks}.
\newblock \bibinfo{journal}{\emph{Inf. Process. Manag.}}  \bibinfo{volume}{56}
  (\bibinfo{year}{2019}), \bibinfo{pages}{1457--1467}.
\newblock


\bibitem[\protect\citeauthoryear{Lukasik, Bontcheva, Cohn, Zubiaga, Liakata,
  and Procter}{Lukasik et~al\mbox{.}}{2019}]%
        {lukasik2016using}
\bibfield{author}{\bibinfo{person}{Michal Lukasik}, \bibinfo{person}{Kalina
  Bontcheva}, \bibinfo{person}{Trevor Cohn}, \bibinfo{person}{Arkaitz Zubiaga},
  \bibinfo{person}{Maria Liakata}, {and} \bibinfo{person}{Rob Procter}.}
  \bibinfo{year}{2019}\natexlab{}.
\newblock \showarticletitle{Gaussian Processes for Rumour Stance Classification
  in Social Media}.
\newblock \bibinfo{journal}{\emph{ACM Trans. Inf. Syst.}} \bibinfo{volume}{37},
  \bibinfo{number}{2} (\bibinfo{year}{2019}).
\newblock


\bibitem[\protect\citeauthoryear{Qazvinian, Rosengren, Radev, and
  Mei}{Qazvinian et~al\mbox{.}}{2011}]%
        {qazvinian2011rumor}
\bibfield{author}{\bibinfo{person}{Vahed Qazvinian}, \bibinfo{person}{Emily
  Rosengren}, \bibinfo{person}{Dragomir Radev}, {and} \bibinfo{person}{Qiaozhu
  Mei}.} \bibinfo{year}{2011}\natexlab{}.
\newblock \showarticletitle{Rumor has it: Identifying misinformation in
  microblogs}. In \bibinfo{booktitle}{\emph{Proceedings of the 2011 Conference
  on Empirical Methods in Natural Language Processing}}.
  \bibinfo{pages}{1589--1599}.
\newblock


\bibitem[\protect\citeauthoryear{Rubanova, Chen, and Duvenaud}{Rubanova
  et~al\mbox{.}}{2019}]%
        {rubanova2019latent}
\bibfield{author}{\bibinfo{person}{Yulia Rubanova}, \bibinfo{person}{Ricky~TQ
  Chen}, {and} \bibinfo{person}{David Duvenaud}.}
  \bibinfo{year}{2019}\natexlab{}.
\newblock \showarticletitle{Latent ODEs for irregularly-sampled time series}.
  In \bibinfo{booktitle}{\emph{Proceedings of the 33rd International Conference
  on Neural Information Processing Systems}}. \bibinfo{pages}{5320--5330}.
\newblock


\bibitem[\protect\citeauthoryear{Schuster and Paliwal}{Schuster and
  Paliwal}{1997}]%
        {schuster1997bidirectional}
\bibfield{author}{\bibinfo{person}{Mike Schuster} {and}
  \bibinfo{person}{Kuldip~K Paliwal}.} \bibinfo{year}{1997}\natexlab{}.
\newblock \showarticletitle{Bidirectional recurrent neural networks}.
\newblock \bibinfo{journal}{\emph{IEEE transactions on Signal Processing}}
  \bibinfo{volume}{45}, \bibinfo{number}{11} (\bibinfo{year}{1997}),
  \bibinfo{pages}{2673--2681}.
\newblock


\bibitem[\protect\citeauthoryear{Tian, Zhang, Wang, and Liu}{Tian
  et~al\mbox{.}}{2020}]%
        {tian2020early}
\bibfield{author}{\bibinfo{person}{Lin Tian}, \bibinfo{person}{Xiuzhen Zhang},
  \bibinfo{person}{Yan Wang}, {and} \bibinfo{person}{Huan Liu}.}
  \bibinfo{year}{2020}\natexlab{}.
\newblock \showarticletitle{Early detection of rumours on Twitter via stance
  transfer learning}. In \bibinfo{booktitle}{\emph{European Conference on
  Information Retrieval}}. Springer, \bibinfo{pages}{575--588}.
\newblock


\bibitem[\protect\citeauthoryear{Zhuang, Dvornek, Li, Tatikonda, Papademetris,
  and Duncan}{Zhuang et~al\mbox{.}}{2020}]%
        {ACA12}
\bibfield{author}{\bibinfo{person}{Juntang Zhuang}, \bibinfo{person}{Nicha
  Dvornek}, \bibinfo{person}{Xiaoxiao Li}, \bibinfo{person}{Sekhar Tatikonda},
  \bibinfo{person}{Xenophon Papademetris}, {and} \bibinfo{person}{James
  Duncan}.} \bibinfo{year}{2020}\natexlab{}.
\newblock \showarticletitle{Adaptive checkpoint adjoint method for gradient
  estimation in neural ode}. In \bibinfo{booktitle}{\emph{International
  Conference on Machine Learning}}. PMLR, \bibinfo{pages}{11639--11649}.
\newblock


\bibitem[\protect\citeauthoryear{Zubiaga, Aker, Bontcheva, Liakata, and
  Procter}{Zubiaga et~al\mbox{.}}{2018a}]%
        {zubiaga2018detection}
\bibfield{author}{\bibinfo{person}{Arkaitz Zubiaga}, \bibinfo{person}{Ahmet
  Aker}, \bibinfo{person}{Kalina Bontcheva}, \bibinfo{person}{Maria Liakata},
  {and} \bibinfo{person}{Rob Procter}.} \bibinfo{year}{2018}\natexlab{a}.
\newblock \showarticletitle{Detection and resolution of rumours in social
  media: A survey}.
\newblock \bibinfo{journal}{\emph{ACM Computing Surveys (CSUR)}}
  \bibinfo{volume}{51}, \bibinfo{number}{2} (\bibinfo{year}{2018}),
  \bibinfo{pages}{1--36}.
\newblock


\bibitem[\protect\citeauthoryear{Zubiaga, Kochkina, Liakata, Procter, and
  Lukasik}{Zubiaga et~al\mbox{.}}{2016}]%
        {zubiaga-etal-2016-stance}
\bibfield{author}{\bibinfo{person}{Arkaitz Zubiaga}, \bibinfo{person}{Elena
  Kochkina}, \bibinfo{person}{Maria Liakata}, \bibinfo{person}{Rob Procter},
  {and} \bibinfo{person}{Michal Lukasik}.} \bibinfo{year}{2016}\natexlab{}.
\newblock \showarticletitle{Stance Classification in Rumours as a Sequential
  Task Exploiting the Tree Structure of Social Media Conversations}. In
  \bibinfo{booktitle}{\emph{Proceedings of {COLING} 2016, the 26th
  International Conference on Computational Linguistics: Technical Papers}}.
  \bibinfo{pages}{2438--2448}.
\newblock


\bibitem[\protect\citeauthoryear{Zubiaga, Kochkina, Liakata, Procter, Lukasik,
  Bontcheva, Cohn, and Augenstein}{Zubiaga et~al\mbox{.}}{2018b}]%
        {ZUBIAGA2018273}
\bibfield{author}{\bibinfo{person}{Arkaitz Zubiaga}, \bibinfo{person}{Elena
  Kochkina}, \bibinfo{person}{Maria Liakata}, \bibinfo{person}{Rob Procter},
  \bibinfo{person}{Michal Lukasik}, \bibinfo{person}{Kalina Bontcheva},
  \bibinfo{person}{Trevor Cohn}, {and} \bibinfo{person}{Isabelle Augenstein}.}
  \bibinfo{year}{2018}\natexlab{b}.
\newblock \showarticletitle{Discourse-aware rumour stance classification in
  social media using sequential classifiers}.
\newblock \bibinfo{journal}{\emph{Information Processing \& Management}}
  \bibinfo{volume}{54}, \bibinfo{number}{2} (\bibinfo{year}{2018}),
  \bibinfo{pages}{273--290}.
\newblock


\end{thebibliography}


\end{document}